\documentclass[11pt]{article}

\usepackage[preprint]{acl}

\usepackage{times}
\usepackage{latexsym}
\usepackage{inconsolata}
\usepackage{tikz}
\usepackage{xcolor,colortbl} % table coloring, place before arydshln
\usepackage{arydshln} 
\usepackage{amsmath}
\usepackage{amssymb} % for \checkmark
\usepackage{multirow}
\usepackage{cleveref}

\usetikzlibrary{positioning, arrows, shapes.multipart}

\usepackage[T1]{fontenc}
\usepackage[utf8]{inputenc}
\usepackage{tipa}

\newcommand{\ipa}[1]{\textipa{#1}}
\newcommand{\wpad}{\textcolor{white}{0}}
\definecolor{brandBlue}{HTML}{648FFF}
\definecolor{brandViolet}{HTML}{785EF0}
\definecolor{brandMagenta}{HTML}{DC267F}
\definecolor{brandOrange}{HTML}{FE6100}
\definecolor{brandYellow}{HTML}{FFB000}
\definecolor{brandBlack}{HTML}{000000}
\definecolor{brandWhite}{HTML}{FFFFFF}
\definecolor{brandGray}{HTML}{3B333D}

\usepackage{microtype}
\usepackage{graphicx}
\usepackage{tabularray}
\usepackage{booktabs}
\usepackage{pifont}% http://ctan.org/pkg/pifont
\usepackage{xspace}

\newcommand{\POWSM}{POWSM\xspace}

\title{\POWSM: A Phonetic Open Whisper-Style Speech Foundation Model}

\author{
 \textbf{Chin-Jou Li$^*$\textsuperscript{1}},
 \textbf{Kalvin Chang$^*$\textsuperscript{2}},
 \textbf{Shikhar Bharadwaj\textsuperscript{1}},
 \textbf{Eunjung Yeo\textsuperscript{3}},
 \textbf{Kwanghee Choi\textsuperscript{3}},
\\
 \textbf{Jian Zhu\textsuperscript{4}},
 \textbf{David R. Mortensen\textsuperscript{1}},
 \textbf{Shinji Watanabe\textsuperscript{1}},
\\
 \textsuperscript{1}Carnegie Mellon University,
 \textsuperscript{2}University of California, Berkeley,
\\
 \textsuperscript{3}University of Texas, Austin,
 \textsuperscript{4}University of British Columbia
\\
}

\begin{document}
\maketitle

\begin{abstract}
Recent advances in spoken language processing have led to substantial progress in phonetic tasks such as automatic speech recognition (ASR), phone recognition (PR), grapheme-to-phoneme conversion (G2P), and phoneme-to-grapheme conversion (P2G). Despite their conceptual similarity, these tasks have largely been studied in isolation, each relying on task-specific architectures and datasets. 
In this paper, we introduce \POWSM (Phonetic Open Whisper-style Speech Model), the first unified framework capable of jointly performing multiple phone-related tasks. 
\POWSM enables seamless conversion between audio, text (graphemes), and phones, opening up new possibilities for universal and low-resource speech processing.
Our model outperforms or matches specialized PR models of similar size (Wav2Vec2Phoneme and ZIPA) while jointly supporting G2P, P2G, and ASR.
% \eunjung{need to change the results to include ASR / PR / other subtasks}
% \POWSM achieves an average 2.6 PFER on in-domain languages, outperforming prior phoneme recognition models such as ZIPA-CR-NS (2.7 PFER), while jointly supporting G2P, P2G, and ASR.
Our training data, code\footnote{\texttt{https://github.com/espnet}} and model\footnote{\texttt{https://huggingface.co/espnet/powsm}} are released to foster open science.

\end{abstract}

\section{Introduction}

% \kwanghee{unify phone (phonetic) or phoneme (phonemic) across the full paper}
% \kwanghee{unify PR/UPR across the full paper}

\begin{figure}[t]
\centering
  \includegraphics[width=\linewidth]{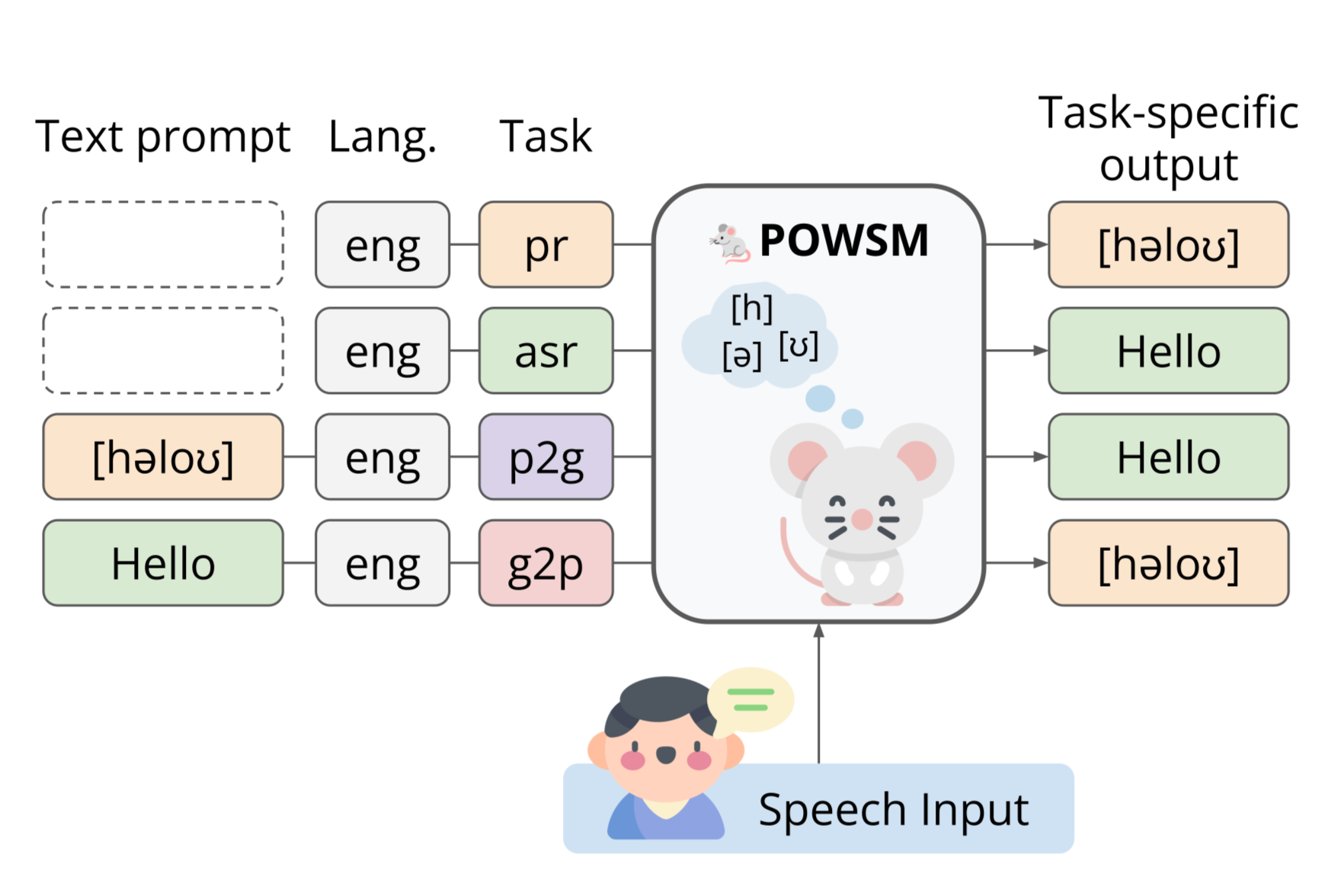}
  \caption{\POWSM is the first phonetic foundation model that can perform four phone-related tasks: Phone Recognition (PR), Automatic Speech Recognition (ASR), audio-guided grapheme-to-phoneme conversion (G2P), and audio-guided phoneme-to-grapheme conversion (P2G).
  }
  \label{fig:overview}
\end{figure}

% Fixed (3.2): \reviewer{Consider including a small qualitative example illustrating cross-task conversion (e.g., audio→phones→text) in the main text}
% Fixed: \reviewer{``phoneme-to-grapheme'' is occasionally written as ``P2G'' without consistent capitalization — ensure consistency.}

% Universal phone recognition (UPR) is the task of transcribing speech into phones, the smallest identifiable unit of sound in speech.
Phones are the smallest units of sound in speech. % distinctive -> phoneme
Unlike graphemes, phones are shared across languages and usually represented using the International Phonetic Alphabet (IPA) \citep{international1999handbook}, a unified transcription standard for all languages.
By providing a consistent representation of speech across languages, phone-level modeling allows fine-grained analysis and cross-lingual generalization, enabling tasks like atypical speech analysis (\textit{e.g.}, L2 speech \cite{li2016mispronunciation,inceoglu2023assessment} and pathological speech \cite{choi-etal-2025-leveraging, li2025towards}), endangered language documentation \cite{he-etal-2024-wav2gloss}, code-switched text-to-speech \cite{9054722}, and cross-lingual transfer in speech-to-text \cite{pratap2024scaling, magoshi25_interspeech}.

% pronunciation training \cite{kheir2025towards}, 
% child speech recognition \cite{dudy2018automatic, lin2024leveraging}, code-switched text-to-speech \cite{9054722}, and cross-lingual transfer in speech-to-text (S2T) \cite{pratap2024scaling, magoshi25_interspeech}.

Four key phone-related tasks underpin phonetic spoken language processing: automatic speech recognition (ASR), phone recognition (PR), grapheme-to-phoneme conversion (G2P), and phoneme-to-grapheme conversion (P2G).
\textit{ASR} learns implicit phonetic representations \cite{belinkov2017analyzing}, while \textit{PR} offers explicit phone-level supervision.
\textit{G2P} and \textit{P2G} bridge orthographic and phonetic spaces. 
Collectively, these tasks interact through shared phonetic representations, each addressing a different aspect of the relationship between audio, phones, phonemes, and graphemes.

Despite their conceptual similarity, these tasks have traditionally been developed in isolation, using task-specific architectures and datasets. 
Such systems are optimized for specific input-output mappings and cannot be easily extended to other phonetic tasks.
This fragmentation has hindered the development of general-purpose models for phonetic processing, necessitating a unified phonetic foundation model that can perform multiple phone-related tasks within a single, general framework for speech processing.

To bridge this gap, we propose \POWSM, a phonetic foundation model capable of performing four core phone-related tasks --- PR, ASR, audio-guided G2P, and audio-guided P2G --- within one unified architecture (\Cref{fig:overview}). 
To construct this framework, we reformulate standard ASR datasets \cite{zhu-etal-2025-zipa} into four task-specific formats, allowing the model to learn consistent mappings across audio, phoneme, and grapheme representations.
In addition, \POWSM adopts an attention-based encoder-decoder (AED) architecture, following the design of large-scale speech foundation models such as Whisper \citep{radford2023robust} and OWSM \citep{peng2023reproducing}.

% Fixed: \reviewer{consider breaking into two sentences for clarity.}
Empirically, \POWSM outperforms previous PR models on both in-domain data and out-of-domain languages and  achieves low-resource ASR performance comparable to web-scale multilingual foundation models. Moreover, it performs speech-grounded P2G and G2P across more than 70 languages.
\POWSM offers a new unified paradigm for phone-level modeling, paving the way for inclusive and globally accessible speech technologies that transcend language boundaries and resource disparities.

To summarize, our main contributions are: 
\begin{itemize}
    \item We release \POWSM, a large-scale foundation model that achieves state-of-the-art PR performance, and is capable of performing multiple fundamental phone-related tasks. Our model enables seamless conversion between speech, text (graphemes/orthography), and phones.
    \item We thoroughly analyze \POWSM to understand the interaction between multiple tasks, architecture components, and losses.
    \item We fully open-source all our data preparation and evaluation scripts, model checkpoint and code to foster open science.
\end{itemize}

\section{Related Work}

\paragraph{Speech foundation models}
Recent speech foundation models such as Whisper \citep{radford2023robust} and OWSM \citep{peng2023reproducing,peng24b_interspeech} have driven progress in large-scale multilingual ASR and speech translation, but they do not explicitly address phoneme recognition or articulatory-level supervision. 
Subsequent work \citep{yusuyin2025whistle,fu2025pac} showed that incorporating phoneme-level objectives improves ASR for low-resource and long-tailed settings, while outputting phonemes as an intermediate benefited speech translation \citep{gállego2025speechtotexttranslationphonemeaugmentedcot}.
\POWSM extends this line of work by being the first open foundation model jointly trained on phone recognition and related tasks, integrating multilinguality, phonetic supervision, and multi-task scalability within one framework.

% \paragraph{Language-specific phone recognition}
\paragraph{Phone recognition}
Prior work in multilingual phone recognition can broadly be categorized into (1) language-specific models \citep{gao21_interspeech} that rely on explicit phoneme \citep{xu22b_interspeech} or allophone inventories \citep{li2020universal} and (2) language-agnostic approaches that aim to generalize across languages without such resources \citep{taguchi23_interspeech,glocker23_interspeech,li2021hierarchical,zhu-etal-2025-zipa}. 
\POWSM follows the latter paradigm as a fully data-driven multilingual model that learns phone representations without predefined phoneme mappings.

WhisperPPT \citep{samir2024efficiently} improved Whisper \citep{radford2023robust}'s performance through data cleaning but remained limited in data coverage and task diversity.
However, Whisper is trained on a closed corpus and could display harmful biases for PR which cannot be fully removed by fine-tuning.
\POWSM is trained from scratch on open datasets.

ZIPA \citep{zhu-etal-2025-zipa} scaled PR to 17,000+ hours of data and 88 languages using a Zipformer \cite{yao2023zipformer} encoder and noisy-student training on 4,000+ languages, achieving state-of-the-art results. To construct its training corpus, ZIPA employed a G2P system to convert large-scale ASR transcriptions into phoneme sequences, effectively repurposing ASR datasets for PR. Building on this idea, \POWSM leverages both the grapheme and the G2P-generated phoneme transcriptions, reformulating them into four task-specific forms: ASR, PR, G2P, and P2G.

\textbf{G2P \& P2G} \POWSM is the first model capable of both audio-guided G2P and audio-guided P2G. 
G2P conversion, sometimes called phonemization in the text-to-speech literature, can be accomplished with pronunciation dictionaries \citep{rudnicky2003cmu}, rules \citep{mortensen2018epitran}, WFSTs \citep{black2001flite}, seq2seq models \citep{zhu2022charsiu-g2p}, or LLMs \citep{qharabagh2025llm}. % to choose between different pronunciations of a word in context
Text-based G2P, however, still cannot handle phonetic variation, enforcing a one-to-one mapping between orthography and transcription.
In contrast, audio-guided G2P can learn to map the different acoustic realizations of a phoneme across varieties of a language to a phone representation \cite{route-etal-2019-multimodal}.
In particular, \citet{mak2025speech} observed a performance improvement by using audio-guided G2P versus text-based G2P alone for Cantonese.
\citet{gao2024g2pu} similarly showed that joint learning of G2P, phone recognition, and forced alignment outperform a G2P teacher model. Similarly, \citet{sun2024acquiring} jointly learned G2P and TTS.
Compared to G2P, P2G conversion is less studied but can be performed with a seq2seq model \citep{lauc2024polyipa} or a finetuned LLM \citep{ma2025llm}.  %on 19 million language-grapheme-phoneme triplets and .
% \kalvin{LLM-based phoneme-to-grapheme for phoneme-based speech recognition}
% To the best of our knowledge, \POWSM is the first to attempt audio-based P2G.

% \kalvin{Acquiring Pronunciation Knowledge from Transcribed Speech Audio via Multi-task Learning: joint learning of G2P and TTS (regressing on the acoustic feature sequence)}

% Improving seq2seq tts frontends with transcribed speech audio ?
% Speech-Based Phonetic Transcript Metrics

% A Survey of Grapheme-to-Phoneme Conversion Methods\\
% Data driven grapheme-to-phoneme representations for a Lexicon-free text-to-speech: neural
% IPA-CHILDES \& G2P+: Feature-Rich Resources for Cross-Lingual Phonology and Phonemic Language Modeling

% \kalvin{what does eSpeak use? \url{https://github.com/espeak-ng/espeak-ng}}

\section{Methodology}
\subsection{Data preparation}
We use IPAPack++ \cite{zhu-etal-2025-zipa} for training. It is an open source corpus of roughly 17,000 hours of multilingual speech with paired orthographic and phonemic transcriptions. We will release all data processing scripts to make \POWSM fully reproducible.

G2P-generated transcriptions have been manually inspected and cleaned. 
% Following \citet{samir2024efficiently}, we remove Interlingua and 10 noisy FLEURS languages. \chinjou{we accidentally included them and the performance is actually better ==}
% ('ga_ie', 'sd_in', 'ar_eg', 'ml_in', 'lo_la', 'da_dk', 'ko_kr', 'ny_mw', 'mn_mn', 'so_so', 'my_mm') 
Utterances longer than 300 phones are filtered out. IPA sequences are normalized to Unicode NFD (Canonical Decomposition); English G2P sequences are further refined with rule-based corrections to fix voice-onset time issues (see Appendix \autoref{sec:appendix-engg2p}).

To prevent IPA tokens from being confused with graphemes, sequences are split into tokens with diacritics and modifiers attached through greedy trie search of PanPhon phone entries \cite{panphon} and enclosed in slashes (e.g., \textipa{/p\super{h}Os@m/} $\rightarrow$ \textipa{/p\super{h}/ /O/ /s/ /@/ /m/}).

\subsection{Multitask data format}
Our model is trained on four tasks: PR, ASR, and audio-guided G2P and P2G. Each utterance is used once per task, with task-specific formatting as illustrated in \Cref{fig:overview}, including a text prompt, language token, task token, and target output. 
We leave the text prompt blank (token \texttt{<na>}) for PR and ASR, and provide graphemes and phones as prompts for G2P and P2G.
For example, for an utterance saying \textit{who is that}, the G2P text prompt is \texttt{"who is that"}, and the target output is \texttt{"<eng><g2p><notimestamps> /h//u//\textipa{I}//z//ð//æ//t/"}.

% The three auxiliary tasks improve training efficiency and model utility. ASR provides orthographic transcriptions, which help regularize the speech-to-text mapping and ground the model in stable phonetic representations. Audio-guided G2P and P2G connect IPA and orthographic representations. They can be seen as context-augmented variants of PR and ASR: G2P predicts orthography given speech and phonemes, while P2G is the inverse. This setup allows the model to leverage both forms, connecting graphemes and IPA symbols. 

\subsection{Training details}
\POWSM adopts an attention-based encoder-decoder (AED) architecture, which flexibly models output sequences and allows the integration of additional tasks.
Specifically, we follow the OWSM v3.1 architecture \cite{peng24b_interspeech}, which employs an E-Branchformer encoder and a Transformer decoder, consistent with the general encoder-decoder structure of Whisper \cite{radford2023robust}. The model is trained from scratch using ESPnet \cite{watanabe2018espnet} with a hybrid CTC/attention loss as in \autoref{eq:loss} \cite{watanabe2017hybridctc}, where we set the ratio $\alpha_\text{ctc}$ to 0.3.
\begin{equation}
\label{eq:loss}\small
    \mathcal{L} = \alpha_\text{ctc} \mathcal{L}_{\text{ctc}} + (1 - \alpha_\text{ctc}) \mathcal{L}_{\text{attention}}
\end{equation}

The encoder operates at the stride size of 40 ms. Training uses a global batch size of 256. Speech inputs are 16kHz and padded to 20 seconds. The vocabulary consists of 40k tokens, including around 6k phone tokens, language and timestamp tokens, and BPE tokens from orthography. 
The model has approximately 350M parameters with 9 layers for both the encoder and decoder and was trained for around 200 GPUs hours on H100s.
Using a CTC loss \citep{graves2006connectionist}, we align the encoder outputs with a simplified version of the phone token sequences. Unlike the decoder outputs, the phones in these sequences are stripped of break (\textipa{/./}, \textipa{/\t*{}/})
and length diacritics (\textipa{/e\textlengthmark/, /e\texthalflength/, /\v{e}/}) to accelerate convergence. Additional details and analyses are provided in \autoref{sec:analysis-ctc-target}. 
The decoder is an autoregressive language model conditioned on a text prompt and attends to the encoder output via cross-attention.

% Skipped: \reviewer{why do we train from scratch instead of initializing from the OWSM 3.1 model?
% a key difference between OWSM and POWSM lies in the encoder representation. OWSM (and most ASR models) uses language-specific BPE targets, whereas our work uses phone entries shared across languages. As discussed in §6.1, mixing phones and orthography in the encoder hindered convergence. Given this and the need for fine-grained control over the phonetic representation space, we chose to train POWSM from scratch to avoid inheriting hidden biases or training artifacts from OWSM.
% } 

\section{Experimental Setup}

\paragraph{Evaluation metric}
We report Phonetic Feature Error Rate (PFER), an edit distance using articulatory features from PanPhon \citep{panphon}, averaged over the number of phones and computed as in \autoref{eq:pfer} for PR. 
Each feature contributes $\frac{1}{24}$ distance unit, while insertion and deletion cost 1 unit. 
The edit distance $D$ grows linearly with the sequence length and has no upper bound.
\begin{equation}
\label{eq:pfer}\small
    \text{PFER} = \frac{1}{\text{\#phone}}\sum_{i=1}^N D(\text{feat}(\text{hyp}_i),\text{feat}(\text{ref}_i))
\end{equation}

Unlike Phone Error Rate (PER), which considers only exact phone matches, or Phone Token Error Rate (PTER), which treats diacritics and modifiers as separate tokens, PFER computes the edit distance in terms of articulatory features---interpretable subphone attributes (\textit{e.g.} voicing)---capturing phonetic similarity in a fine-grained fashion. 
Previous studies \cite{taguchi23_interspeech, zhu-etal-2025-zipa} define PFER as the mean articulatory feature edit distance over the evaluation set. In contrast, we normalize it by the number of phones in the reference transcription to measure the proportion of feature errors per phone.

\paragraph{Decoding hyperparameters}
We use a CTC weight (denoted as \texttt{ctc}) of 0.3 and a beam size (denoted as \texttt{beam}) of 3 during decoding for all reported numbers unless specified. Further details on the choice of hyperparameters are discussed in \autoref{sec:analysis-ctc-weight}.

\paragraph{Evaluation datasets}
For unseen languages, we evaluate on three datasets: DoReCo \cite{paschen2020doreco}, VoxAngeles \cite{chodroff2024voxangeles}, and Tusom2021 \cite{mortensen2021tusom2021}.
% \kalvin{add details about datasets}
DoReCo is a dataset of 50+ languages (with broad transcriptions) intended for documentation of small or endangered languages; we use a 45-language subset.\footnote{
% Fixed: \reviewer{why 45 lang subset?}
We use the same DoReCo subset as \citet{zhu-etal-2025-zipa,zhu-etal-2024-taste}, listed in \autoref{sec:appendix-doreco}. They removed languages mostly due to licensing issues, while others were not accessible during dataset creation.
}
VoxAngeles \citep{chodroff2024voxangeles} is a postprocessed version of the UCLA Phonetics Lab Archive \citep{ucla2009} containing 95 languages.
Tusom is a low-data Tangkhulic language of India not included in the training data. Tusom2021 \citep{mortensen2021tusom2021} consists of narrow phonetic transcriptions (unlike the broad transcriptions from G2P on which \POWSM was trained) of individual Tusom words.
We removed the tones.
% \kalvin{how narrow are doreco and voxangeles}

We also test on five datasets on varieties of English: the Buckeye Corpus \cite{pitt2005buckeye} and DoReCo South-England represent dialectal variation, while L2-ARCTIC \cite{zhao2018l2arctic}, EpaDB \cite{vidal2019epadb}, and SpeechOcean762 \cite{zhang2021speechocean762} contain L2 speakers.
% For L2-ARCTIC we only use the "scripted" subset with shorter utterance durations.
For L2-ARCTIC, we used the manually annotated phoneme transcriptions (which \citet{zhu-etal-2025-zipa} termed \textit{L2-Perceived}) rather than G2P dictionary-based transcriptions.  % \kalvin{technically they come from forced alignment but forced alignment was trained with pronouncing dicts}
% https://psi.engr.tamu.edu/l2-arctic-corpus/
% https://huggingface.co/datasets/KoelLabs/L2Arctic
The manual transcriptions reflect what the speaker actually said, whereas the dictionary-based version enforces a single pronunciation variant.\footnote{For instance, ``crayon'' in American English can be pronounced as 
% \ipa{/ˈkɹæn/}, \ipa{/ˈkɹej.ɔn/}, or \ipa{/ˈkɹej.ɒn/} 
\textipa{/\textprimstress k\*r\ae n/}, \textipa{/\textprimstress k\*rej.On/}, or \textipa{/\textprimstress k\*rej.6n/} \citep{vaux2003harvard} (among others), but the CMU Pronouncing Dictionary \citep{rudnicky2003cmu} only lists one.}
Manual inspection by a trained phonologist further showed the L2-ARCTIC transcriptions to be of extremely poor quality.
For the five aforementioned datasets, we use preprocessed datasets from \citet{zhu-etal-2025-zipa}\footnote{\texttt{https://huggingface.co/anyspeech}} and Koel Labs\footnote{texttt{https://huggingface.co/KoelLabs}} for better transcription quality. 
%\drm{Something is wrong with this sentence. It should also be mentioned that manual inspection showed the L2-ARCTIC transcriptions to be of extremely poor quality.} \chinjou{we are switching to l2-arctic-perceived, which would be better}
% \shikhar{Including long samples from suitcase is possible but we need to devise a way to handle long audio which is beyond the scope of this particular work}

We then evaluated our model on in-domain data from IPAPack++, the dataset seen during training. We followed \citet{zhu-etal-2025-zipa} in using LibriSpeech for English, AISHELL for Mandarin, and MLS for European languages, and additionally evaluated on IISc-MILE Tamil \cite{a2022subworddictionarylearningsegmentation} for Tamil and KSC \cite{khassanov-2021-crowdsourced-ksc} for Kazakh.

For ASR and P2G, we evaluate with FLEURS.

See \autoref{tab:test-data-stats} for more details about our evaluation datasets.

\begin{table}[h]
\vspace{-2mm}
\centering
\small
\setlength{\tabcolsep}{4pt} % Adjust column spacing
\renewcommand{\arraystretch}{1.2}
\resizebox{0.95\columnwidth}{!}{
\begin{tabular}{c c c c c c}
\toprule
\rowcolor{gray!10}
\addlinespace[0.10cm]
\multicolumn{6}{l}{\textbf{PR (In-domain)}}\\
\texttt{eng} & \texttt{deu} & \texttt{nld} & \texttt{fra} & \texttt{ita} & \texttt{spa} \\ 
10.58 & 14.27 & 12.76 & 10.07 & 5.27 & 10.00\\
\texttt{por} & \texttt{pol} & \texttt{tam} & \texttt{kaz} & \texttt{cmn}\\
3.74 & 2.14 & 16.58 & 7.07 & 10.02 \\
\addlinespace[0.10cm]
\rowcolor{gray!10}
\addlinespace[0.10cm]
\multicolumn{6}{l}{\textbf{PR (Out-of-domain: Unseen languages)}}\\
% \multicolumn{2}{c}{DoReCo} &  \multicolumn{2}{c}{VoxAngeles} & \multicolumn{2}{c}{Tusom2021}\\
% \multicolumn{2}{c}{19.18} & \multicolumn{2}{c}{1.58} & \multicolumn{2}{c}{1.16}\\
DoReCo &  VoxA. & Tusom.\\
19.18 & 1.58 & 1.16\\
\addlinespace[0.10cm]
\rowcolor{gray!10}
\addlinespace[0.10cm]
\multicolumn{6}{l}{\textbf{PR (Out-of-domain: Language variation)}}\\
Buckeye & DRC-SE & L2-ARC & EpaDB & SO762\\
7.88 & 0.77 & 3.66 & 2.74 & 2.32\\ 
\midrule
\midrule
\rowcolor{gray!10}
\multicolumn{6}{l}{\textbf{ASR (FLEURS)}}\\
\texttt{afr} & \texttt{orm} & \texttt{aze}  & \texttt{pan} &\texttt{tgk}  & \texttt{mkd} \\
0.66 & 0.13 & 2.37 & 1.48 & 1.96 & 2.45 \\
\texttt{bos} & \texttt{slv} \\
2.45 & 1.76\\
\bottomrule
  \end{tabular}
}
 \caption{Duration of the test sets for different tasks (in hours). Abbreviated datasets (in order): VoxAngeles, Tusom2021, DoReCo South-England, L2-ARCTIC, SpeechOcean762.}
\label{tab:test-data-stats}
\vspace{-2mm}
\end{table}

\paragraph{Baselines}
We evaluate all PR baselines without further training with IPAPack++. See Appendix \autoref{sec:appendix-baseline} for more details about training data and language coverage.
Allosaurus \cite{li2020universal,li2021hierarchical} uses a phone-level CTC to train a language-agnostic model and applies language-specific allophone-to-phoneme mappings. 
Wav2Vec2Phoneme \citep{xu22b_interspeech}, MultIPA \cite{taguchi23_interspeech} and Allophant \cite{glocker23_interspeech} fine-tune XLS-R \citep{babu22_interspeech} with different objectives: Wav2Vec2Phoneme maps unseen phonemes using articulatory features, MultIPA leverages high-quality G2P data from seven languages, while Allophant decomposes phones into articulatory features and applies CTC losses for each. 
ZIPA \cite{zhu-etal-2025-zipa} trains ZipFormer \citep{yao2023zipformer} from scratch on IPAPack++ using CR-CTC and also provides a variant trained with additional pseudo-labeled data (``ZIPA-CR-NS-Large'').

For ASR, we compare \POWSM with two series of models: OWSM \cite{peng25c_interspeech} and OWLS \cite{chen2025owls}. We select OWSM-CTC v4 because it is the best-performing model in the series, featuring an encoder-CTC architecture that supports ASR, ST, and LID. For OWLS, we include models with comparable parameter sizes.

\section{Results}
We find that \POWSM achieves comparable or better performance on PR and ASR than competitive baselines, particularly in low-resource settings; audio-G2P and P2G are discussed in \autoref{sec:analysis-task-lang}. These results suggest that including PR as a pre-training task improves representation generality and reduces the data required to map acoustics to text tokens.
In contrast, we find no clear evidence that ASR benefits PR under the current setup (\autoref{sec:appendix-asr-perf}, \autoref{sec:appendix-multitask}).
% Fixed: \reviewer{paste our response here, about multitask. also point to appendix about ASR and multitask}
% Fixed: \reviewer{I believe it is necessary to include results on common ASR benchmarks (e.g., LibriSpeech) with more comprehensive metrics, as ASR is also one of the main training tasks. Results solely on PR tasks and low-resource ASR tasks are, in my opinion, insufficient to fully demonstrate the model's capabilities.} 
% Fixed: \reviewer{updated results after fixing librispeech text normalization issue} \chinjou{appendix}

\subsection{Multi-task performance}
Results on the in-domain test sets are presented in \autoref{tab:main-seen} and \autoref{tab:main-asr}. We provide further discussion of G2P and P2G in \autoref{sec:analysis-task-lang}.

\paragraph{\POWSM excels at in-domain phone recognition} From \autoref{tab:main-seen}, we see that \POWSM achieves the lowest average PFER in phone recognition, due to the strong language modeling capability of the decoder.
We hypothesize that our English data cleaning (Appendix \autoref{sec:appendix-engg2p}) may have negatively affected the PFER for Germanic languages due to a mismatch between training and test data. 
Nevertheless, our approach fills this gap by achieving strong performance on other languages, outperforming models trained on larger datasets.

\begin{table*}[t]
\vspace{-2mm}
\centering
\resizebox{0.95\textwidth}{!}{
\begin{tabular}{lc rrr rrrr rrrr r}
\toprule
Model & Param. & \texttt{eng } & \texttt{deu } & \texttt{nld } & \texttt{fra } & \texttt{ita } & \texttt{spa } & \texttt{por } & \texttt{pol } & \texttt{tam } & \texttt{kaz } & \texttt{cmn } & Avg.\\
\midrule
Allosaurus & 11M & 6.89 & 17.67 & 19.19 & 20.91 & 19.02 & 4.82 & 19.61 & 21.21 & 12.01 & 20.90 & 15.28 & 16.14\\
Allophant & 300M & 10.26 & 9.37 & 18.39 & 18.83 & 7.82 & 17.37 & 15.44 & 7.90 & 19.32 & --- & --- & ---\\
%W2V2P-lv-60-ft & 300M & 5.93 & 11.39 & 16.13 & 19.08 & 7.09 & 5.43 & 14.53 & 9.98 & 14.50 & 16.75 & 16.09 & 12.45\\
Wav2Vec2Phoneme & 300M & 7.70 & 7.89 & 12.31 & 17.73 & 6.10 & 3.67 & 11.65 & 9.57 & 15.63 & 15.30 & 14.66 & 11.11\\
MultIPA & 300M & 15.81 & 16.28 & 18.97 & 20.19 & 7.20 & 6.99 & 15.04 & 2.63 & 10.54 & 17.71 & 21.10 & 13.86\\
ZIPA-CR-Large & 300M & 1.63 & 3.32 & 3.03 & 3.23 & 3.24 & 1.98 & 4.01 & 4.33 & 4.59 & 2.31 & 1.25 & 2.99\\
ZIPA-CR-NS-Large & 300M & \textbf{1.40}& \textbf{3.17}& \textbf{2.83} & \textbf{2.92} & 3.22 & 1.53 & 3.40 & 4.31 & 4.15 & \textbf{1.87} & \textbf{0.90} & 2.70\\
\midrule
POWSM & 350M & 2.85 & 3.37 & 5.14 & 3.27 & \textbf{1.81} & \textbf{1.21} & \textbf{2.90} & \textbf{1.36} & \textbf{3.56} & 2.25 & 1.15 & \textbf{2.62}\\
%POWSM medium & 1B\\
\bottomrule
\end{tabular}
}
\caption{PFER ($\downarrow$) on the in-domain dataset, IPAPack++. Languages not supported by Allophant are left blank. Some languages were not seen by MultiIPA. \textbf{Bold} indicates the best performance.}
\label{tab:main-seen}
\vspace{-2mm}
\end{table*}

\paragraph{\POWSM is comparable with web-scale ASR models on low-resource languages}
We hypothesize that pre-training with phone recognition benefits low-resource ASR \cite{yusuyin2025whistle}. 
To choose low-resource languages, we selected languages in IPAPack++ with less than 8 hours of speech in FLEURS to serve as the test set. 
See \autoref{sec:appendix-asr-splits} for details on the amount of data used by different models.
For a fair comparison with other multilingual ASR baselines without language-specific components, we use the same decoding hyperparameters \texttt{ctc=0.0, beam=1}.

As shown in \autoref{tab:main-asr}, \POWSM (\texttt{POWSM 0.35B, ASR}) is often comparable to models of similar size trained on web-scale data for ASR (\texttt{OWLS 0.5B}).
Incorporating phones obtained from PR as text prompts (PR-P2G) significantly decreases WER, making it comparable to or even better than these models.
When using gold phone labels for P2G (see analysis in \autoref{sec:analysis-p2g-good}), \POWSM outperforms other ASR models by a large margin in most cases.

\begin{table}
\vspace{-2mm}
\centering
\small
\setlength{\tabcolsep}{4pt} % Adjust column spacing
\renewcommand{\arraystretch}{1.4}
\resizebox{\columnwidth}{!}{
\begin{tabular}{l cc c cc ccc}
\toprule
% use numbers reported in OWSM v4
& \multicolumn{2}{c}{Afroasiatic} & Turkic & \multicolumn{2}{c}{Indo-Iranian} & \multicolumn{3}{c}{Balto-Slavic}\\
Model & \texttt{afr} & \texttt{orm} & \texttt{aze}  & \texttt{pan} & \texttt{tgk}  & \texttt{mkd} & \texttt{bos} & \texttt{slv}\\
\midrule
%\rowcolor{gray!10}
%MMS 1B-all & 24.4 & 65.3 & 23.3 & 33.7 & 14.8 & 8.1 & 13.4 & 14.4\\
OWLS 0.5B  & 102.3 & \textbf{89.0} & 77.7 & \underline{59.3} & 60.4 & 54.2 & 59.3 & \underline{58.6}\\
OWLS 1B  & 95.7 & 102.4 & \underline{67.5} & \textbf{50.0} & \textbf{50.7} & \textbf{46.2} & \textbf{50.0} & \textbf{52.8}\\
OWSM-CTC v4 1B & \textbf{67.5} & \underline{92.7} & 71.2 & 88.7 & 57.6 & \underline{51.2} & \underline{51.3} & 60.4\\
\midrule
POWSM 0.35B, ASR & 86.2 & 125.3 & 67.7 & 83.1 & 62.8 & 56.0 & 56.5 & 64.5\\
%POWSM 1B, ASR &  &  &  &  &  &  &  & \\
POWSM 0.35B, PR-P2G & \underline{68.8} & 93.0 & \textbf{66.7} & 72.8 & \underline{51.0} & \underline{48.6} & 56.9 & 63.9\\
%POWSM 1B, PR-P2G \\
\bottomrule
\end{tabular}
}
\caption{WER ($\downarrow$) of ASR and PR-P2G on low-resource languages. PR-P2G uses phones predicted by PR as text prompts instead of gold phones. \textbf{Bold} indicates the best performance, and \underline{underline} indicates the second-best.}
\label{tab:main-asr}
\vspace{-2mm}
\end{table}

\subsection{\POWSM generalizes well to unseen languages} % PR on out-of-domain data
\autoref{tab:main-unseen} reports PFER on datasets with unseen languages and language variation.
Results indicate that \POWSM achieves strong performance across these datasets, and handles both dialectal and L2 variation effectively.
Notably, our method outperforms ZIPA trained on the same data and even exceeds ZIPA trained with extra pseudo-labeled data, achieving the best results on unseen languages while performing three additional tasks.
While \POWSM lags behind Wav2Vec2Phoneme on socio-phonetic variations, we attribute this to its self-supervised learning with over 60k hours of speech (from wav2vec 2.0 \citep{baevski2020wav2vec}) prior to the supervised learning stage.

\begin{table*}[t]
\vspace{-2mm}
\centering
\resizebox{0.95\textwidth}{!}{
\begin{tabular}{lc cccc|ccrrcc}
\toprule
& &\multicolumn{4}{c}{Unseen Languages} & \multicolumn{6}{c}{Language Variation}\\
Model & Param. & {DoReCo} & {VoxAngeles} & {Tusom2021} & Avg. & {Buckeye} & {DRC-SE} & {L2-ARC} & {EpaDB} & {SO762} & Avg. \\
\midrule
Allosaurus & 11M & 24.71 & 30.84 & 42.02 & 32.52 & 15.24 & 25.36 & 13.39 & 19.33 & 21.61 & 18.99\\
Allophant & 300M & --- & --- & --- & --- & 16.05 & 24.13 & 11.91 & 14.38 & 18.28 & 16.95\\
%W2V2P-lv-60-ft & 300M & 17.76 & 15.69 & 35.30 & 22.92 & 12.08 & 19.34 & 7.74 & 10.11 & 15.21 & 12.90\\
Wav2Vec2Phoneme & 300M & 17.25 & \textbf{13.88} & 31.92 & 21.02 & 12.50 & 18.57 & 9.86 & \textbf{9.90} & \textbf{13.60} & \textbf{12.89}\\
MultIPA & 300M & 18.28 & 15.23 & 30.53 & 21.35 & 18.69 & 23.31 & 15.52 & 15.64 & 21.34 & 18.90\\
ZIPA-CR-Large & 300M & 17.99 & 16.95 & 23.68 & 19.54 & \textbf{12.04} & 17.89 & 9.74 & 17.38 & 15.58 & 14.53\\
ZIPA-CR-NS-Large & 300M & \textbf{16.82} & 17.14 & 23.08 & 19.01 & 12.05 & \textbf{17.12} & \textbf{9.69} & 14.63 & 18.20 & 14.34\\
\midrule
POWSM & 350M  & 17.06 & 17.11 & \textbf{21.96} & \textbf{18.71} & 12.63 & 18.33 & 11.32 & 11.86 & 17.84 & 14.40\\
%POWSM medium & 1B\\
\bottomrule
\end{tabular}
}
\caption{PFER ($\downarrow$) on out-of-domain data. ``DRC-SE'' stands for DoReCo South-England; ``L2-ARC'' stands for L2-ARCTIC; ``SO762'' stands for SpeechOcean762. Unseen language datasets include languages not supported by Allophant; therefore, we do not report results for these datasets.}
\label{tab:main-unseen}
%\vspace{-4mm}
\end{table*}

\newcommand{\cmark}{\ding{51}}%
\newcommand{\xmark}{\ding{55}}%
\section{Analysis}

In this section, we analyze how \POWSM works, focusing on the phonetic-aware encoder and task- and language-specific tokens, which are the defining features of the model.

\subsection{Behavior of the speech encoder}

\paragraph{The CTC encoder prefers fine-grained phones without suprasegmentals}
\label{sec:analysis-ctc-target}
We observed that mixing phones and orthography as encoder targets hindered training, because the same speech input would have different encoder CTC targets for different tasks. 
Therefore, we used phones as encoder targets, encouraging general representations of sounds to be shared across languages.

To determine the most effective unit for the CTC encoder, we fix the decoder vocabulary to PanPhon phones and compared four encoder targets: (1) Unicode code points vs. PanPhon, and (2) sequences with vs. without suprasegmentals (length and break marks). Unicode code points offer simplicity and a smaller vocabulary but split phones into unnatural units (e.g. \textipa{/p\super{h}/}) and increase sequence length, while PanPhon represents each phone-diacritic combination as a unit (e.g. \textipa{/p\super{h}/}), yielding a more natural monotonic sequence at the expense of sparsity and potential out-of-vocabulary issues. Suprasegmentals such as \textipa{/\textlengthmark/}, though phonemic in many languages, confuse PR models \citep{zhu-etal-2025-zipa}.

\begin{figure}[t]
    \centering
    \includegraphics[width=\linewidth]{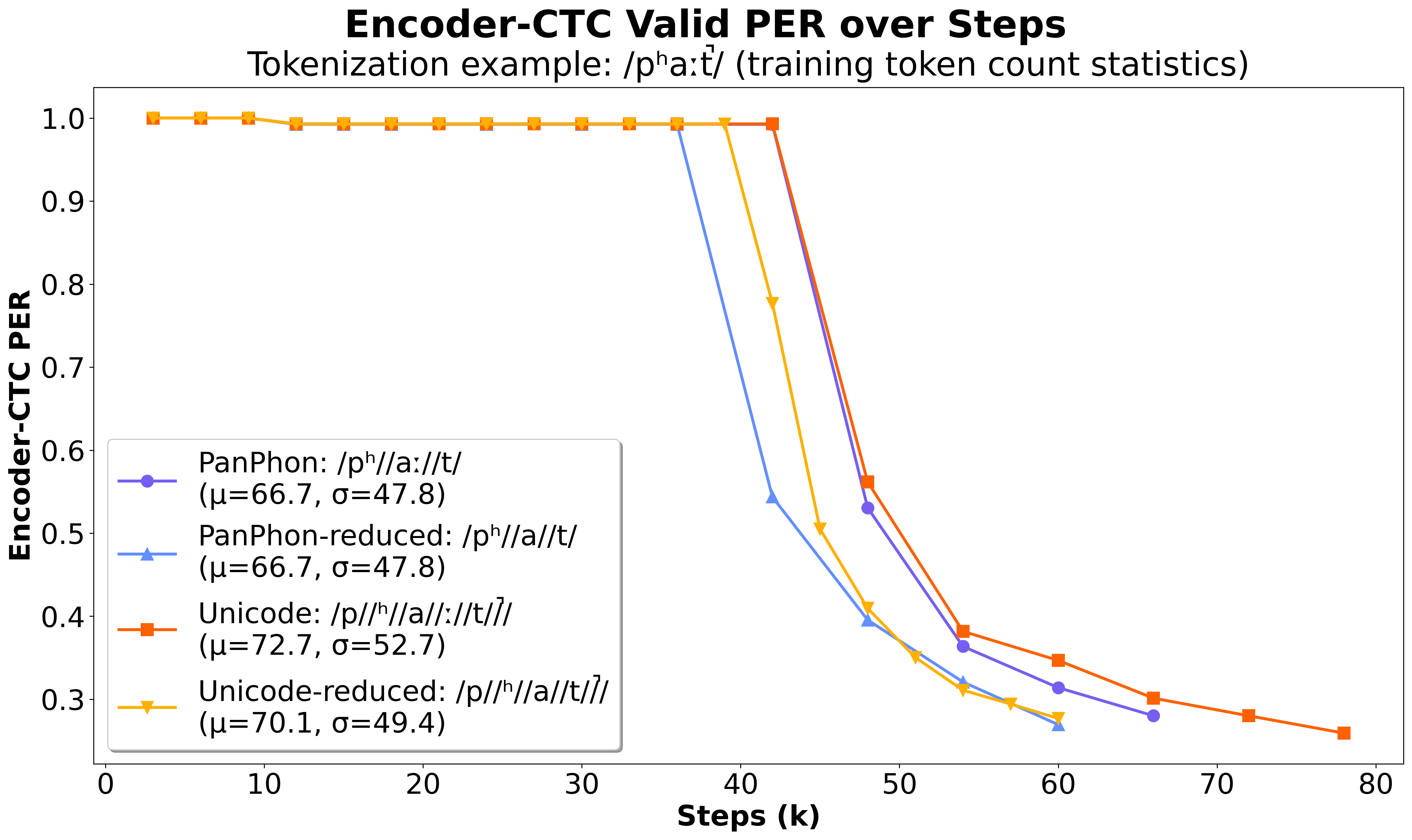}
    \caption{Validation PER of encoder-CTC during training. Removing suprasegmentals from the training target of encoder accelerates convergence.
    % Fixed: \reviewer{Please enlarge Figure 2 and improve its readability (font size, caption clarity).}
    }
    \label{fig:ctc-vocab}
\end{figure}

We run small-scale experiments on a 1k-hour subset of the multi-task data (250 hours of speech repeated across four tasks).
We use the validation CER of the encoder-CTC output as a proxy for training efficiency. 
An earlier drop indicates that the encoder is learning a useful alignment early, which improves representations fed into the decoder and accelerates overall convergence. 
% fast is ambiguous (could also mean steep convergence)
In \autoref{fig:ctc-vocab}, PanPhon tokenization without suprasegmentals shows the earliest drop, suggesting that alignment with decoder units aids training, while
collapsing suprasegmental distinctions for CTC reduces confusion.

\paragraph{Increased encoder weights benefit PR on out-of-domain data}
\label{sec:analysis-ctc-weight}
As in other encoder-decoder models \cite{gong23d_interspeech,  radford2023robust}, we expect the encoder of \POWSM to capture more general acoustic patterns, while the decoder handles language and task-specific output formats. Therefore, 
% instead of ignoring the encoder output as in most S2T tasks, 
we investigate whether emphasizing the encoder more during different stages of model development affects performance.
To balance data diversity with inference compute costs, we selected two smaller datasets from each category with distinct characteristics.

As shown in \autoref{tab:ctc-weight}, higher CTC decoding weights improve PR performance on out-of-domain data but degrade it on in-domain data, as expected.
This echoes \citet{zhu-etal-2025-zipa}'s finding that RNN-T \citep{graves2014towards}, an encoder-only speech-to-text model with an autoregressive text prediction network, hurts generalization to unseen patterns of phones.
% Jian: ``learning the causal dependencies between phones can also hurt the multilingual generalizability, as unseen languages might have a different phonological structure''
We hypothesize that the decoder is performing implicit phonotactic language modeling and ``smooths'' phonetic variation towards more probabilistically likely phone patterns, as \citet{zhu-etal-2025-zipa} described.

Next, we examine whether focusing more on the CTC loss through training widens this gap in performance between in-domain and out-of-domain data.
We find that fine-tuning with a higher CTC loss weight $\alpha_\text{ctc}$ after convergence does not improve out-of-domain performance and can even degrade it. Randomly varying $\alpha_\text{ctc}$ for each batch also shows no improvement.
In contrast, training with a higher $\alpha_\text{ctc}$ from the start benefits the out-of-domain distribution, achieving the lowest PFER on unseen languages with greedy decoding, while the PFER on in-domain data is comparatively higher.
These results suggest that assigning a higher weight to the encoder during training and inference improves PR, highlighting a common trade-off between in-domain performance and generalization.

\begin{table}[h]
\vspace{-2mm}
\centering
\small
\setlength{\tabcolsep}{4pt} % Adjust column spacing
\renewcommand{\arraystretch}{1.2}
\resizebox{\columnwidth}{!}{
\begin{tabular}{l cc| cccc}
\toprule
& \multicolumn{2}{c}{In-domain} & \multicolumn{4}{c}{Out-of-domain}\\
Setup & \texttt{ita} & \texttt{ pol } & VoxA. & Tusom. & DRC-SE & EpaDB\\
\midrule
\multicolumn{3}{l|}{\textbf{Decoding}}\\
\texttt{ctc=0.3} & \textbf{1.66} & \textbf{1.36} & \textbf{17.58} & 33.52 & 18.21 & 11.88\\
\texttt{ctc=0.7} & 1.97 & 1.37 & 17.92 & 24.29 & 18.05 & 11.82\\
\texttt{ctc=0.9} &2.05 & 1.38 & 19.27 & \textbf{22.94} & \textbf{17.67} & \textbf{11.80}\\
\multicolumn{3}{l|}{\textbf{Pre-training / Fine-tuning}}\\
$\alpha_\text{ctc}$=0.3 & \textbf{1.81} & 1.60 & 16.02 & 22.47 & 18.59 & 11.66\\
$\rightarrow$ Ft, $\alpha_\text{ctc}$=0.5& 1.94 & \textbf{1.53} & 17.78 & 23.72 & 18.73 & 11.82\\
%\hspace{1em}$\rightarrow$Ft: $\alpha_\text{ctc}$=0.7& 1.78 & 1.48 & 26.34 & 38.36 & 18.68 & 11.38\\
$\alpha_\text{ctc}$=0.7 & 1.96 & 1.65 & \textbf{15.40} & \textbf{22.10} & 18.50 & 11.62 \\
$\rightarrow$ Ft, $\alpha_\text{ctc}$=0.5& 2.01 & 1.57 & 16.21 & 22.93 & \textbf{18.41} & 11.47 \\

$\alpha_\text{ctc}$=$U$(0.1,0.9) & 1.95 & 1.62 & 19.29 & 25.11 & 18.92 & \textbf{11.33} \\
\bottomrule
  \end{tabular}
  }
\caption{PFER ($\downarrow$) for different CTC weight settings. ``Ft'' denotes fine-tuning for 5 epochs from the checkpoint above. VoxAngeles and Tusom2021 are abbreviated. 
Pre-training and fine-tuning rows use \texttt{ctc=0.3}. All setups use \texttt{beam=1}.
}

\label{tab:ctc-weight}
\end{table}

%%%%%%%%%%%%%%%%%%%%%%%%%%%%%%%%%%%%%%%%%%%%%%%%%%%%%%%%%%%%%%%%%%%%%%%%%%%%%%%%%%%

\begin{table}[h]
\vspace{-2mm}
\centering
\small
\setlength{\tabcolsep}{4pt} % Adjust column spacing
\renewcommand{\arraystretch}{1.4}
\resizebox{\columnwidth}{!}{
\begin{tabular}{l c c c}
\toprule
% Remove EpaDB
Task & Buckeye & Example \\
\midrule
\rowcolor{gray!10}
\multicolumn{2}{l}{ASR Transcription} & any holidays at all they just kind of ignore\\
\rowcolor{gray!10}
% \multicolumn{2}{l}{Phonetic transcription} & \ipa{/ɛnihɑlʌdeɪsɛɾɑl\textcolor{blue}{soʊ}ðeɪdʒʌstkɑɾ̃ʌvɪgnɔɹ/} \\
\multicolumn{2}{l}{Phonetic transcription} & \textipa{/EnihAl2deIsERAl\textcolor{blue}{soU}DeIdZ2stkA\~r2vIgnO\*r/} \\
\midrule
% PR & 12.63  & \ipa{/ɛ̃nihɑlədeɪzætɔl\textcolor{blue}{soʊ}ðeɪtʃʌstkʰæ̃nəvɪɡnɔɹ/}\\
PR & 12.63  & \textipa{/\~EnihAl@deIz\ae tOl\textcolor{blue}{soU}ðeItS2stk\textsuperscript{h}\~\ae n@vIgnO\*r/}\\
% G2P (speech) & 12.71 & \textipa{/ɛ̃nihɑlədeɪzætɔl\textcolor{blue}{soU}ðeɪtʃʌstkʰæ̃nəvɪɡnɔɹ/}\\
G2P (speech) & 12.71 & \textipa{/\~EnihAl@deIz\ae tOl\textcolor{blue}{soU}ðeItS2stk\textsuperscript{h} \~\ae n@vIgnO\*r/}\\
% G2P (both) & 16.38 & \ipa{/ɛ̃nihɑlədeɪzætɔlðeɪtʃʌstkʰ\textcolor{orange}{ɪ̃}ndəvɪɡnɔɹ/}\\
G2P (both) & 16.38 & \textipa{/\~EnihAl@deIz\ae tOlðeItS2stk\textsuperscript{h}\textcolor{orange}{\~I}nd@vIgnO\*r/}\\
% G2P (text prompt) & 23.44 & \ipa{/\textcolor{orange}{aɪhoʊl̴}d\textcolor{orange}{a}ɪzætɔl̴ðeɪtʃɪstkʰ\textcolor{orange}{ɪ̃}ndəvɪɡn\textcolor{orange}{ɜ˞}/}\\
G2P (text prompt) & 23.44 & \textipa{/\textcolor{orange}{aIhoU\textltilde}d\textcolor{orange}{a}Iz\ae tO\textltildeðeItSIst\textsuperscript{h}\textcolor{orange}{\~I}nd@vIgn\textcolor{orange}{\textrhookrevepsilon}/}\\

\bottomrule
  \end{tabular}
  }
\caption{Comparing G2P with different available modalities with PFER ($\downarrow$). \textcolor{blue}{Blue} for correctly capturing mispronounced parts (\textipa{/soU/}), \textcolor{orange}{orange} for error compared to other examples.}
\label{tab:task-token-g2p}
\end{table}

\subsection{Inspecting Task and Language Tokens}
\label{sec:analysis-task-lang}

\paragraph{Speech-guided G2P preserves phonetic variation; text prompts normalize it} % Speech-text interaction in G2P
To better understand how \POWSM integrates speech and text prompts, we analyze the relative influence of speech and text prompts in its G2P behavior.
We vary the G2P conditions from purely speech-based to purely text-based, as shown in \autoref{tab:task-token-g2p}, and evaluate the model on the Buckeye dataset.
When only speech is provided, the performance is comparable to the PR setting, which differs only in the task token.
Adding both speech and text prompts (the standard G2P setup) leads to degraded performance, with output showing standardized pronunciations.
When the model relies solely on the text prompt, performance drops sharply and pronunciations become highly standardized as expected (just as \citet{zhu-etal-2025-zipa} reported).
% Thus text prompting normalizes variation, while speech-guided G2P normalizes 

In other words, \POWSM G2P responds to speech and text signals to controllably mediate between narrow and broad transcription.
% The above observation further affirms \citet{zhu-etal-2025-zipa}'s finding that language variation tends to be normalized.
In the multi-task setup, this effect may be stronger because the model is trained with G2P, which could bias it toward more standardized forms.
% \eunjung{isn't \POWSM G2P (both)? The example shows that POWSM is not responding to speech (not capturing sou)}\chinjou{oh no. but maybe it's okay, because it's G2P...? PR and audio-only G2P is responding}

\paragraph{Audio-P2G effectively handles low-resource languages}
\label{sec:analysis-p2g-good}
We compare several P2G setups on the same set of low-resource languages from FLEURS, listed in \autoref{tab:analysis-task-p2g}. P2G significantly outperforms ASR, suggesting that it effectively leverages the provided phone context. 
However, since P2G uses gold phone labels, this comparison is not entirely fair. We therefore tested PR followed by P2G (PR-P2G), and found that performance improved for some languages but not for others. 
Error propagation does not explain this variation in performance, as PFER trends from PR differ from the observed performance drops. Yet the PFER pattern aligns closely with ASR results, suggesting that phonotactic similarity to high-resource languages may play a role.

To test this, we run P2G with the language code set to English and post-process the output to match Cyrillic or Gurmukhi transcriptions with online conversion tools for certain languages.\footnote{\url{https://www.lexilogos.com} for Macedonian and Tajik; \url{https://punjabi.indiatyping.com} for Panjabi.}
This approach often outperforms ASR and sometimes approaches P2G's performance, indicating that P2G also relies heavily on speech input.
Languages with either comparibly low or high PFER did not benefit from this transliteration approach, possibly because the model already handled them well or had not yet learned them sufficiently. 
This finding suggests a direction for further investigation in low-resource ASR.

\begin{table}
\vspace{-2mm}
\centering
\small
\setlength{\tabcolsep}{4pt} % Adjust column spacing
\renewcommand{\arraystretch}{1.4}
\resizebox{\columnwidth}{!}{
\begin{tabular}{l cc c cc ccc}
\toprule
% use numbers reported in OWSM v4
& \multicolumn{2}{c}{Afroasiatic} & Turkic & \multicolumn{2}{c}{Indo-Iranian} & \multicolumn{3}{c}{Balto-Slavic}\\
Task & \texttt{afr} & \texttt{orm} & \texttt{aze}  & \texttt{pan} & \texttt{tgk}  & \texttt{mkd} & \texttt{bos} & \texttt{slv}\\
\midrule
Best ASR & 67.5 & \underline{89.0} & 67.5 & \textbf{50.0} & 50.7 & \underline{46.2} & 50.0 & \underline{52.8}\\
\midrule
ASR & 86.2 & 125.3 & 67.7 & 83.1 & 62.8 & 56.0 & 56.5 & 64.5\\
P2G & \textbf{55.9} & \textbf{88.0} & \textbf{37.1} & \underline{52.6} & \textbf{31.8} & \textbf{36.9} & \textbf{32.3} & \textbf{40.3}\\
P2G, \texttt{lang=<eng>} & \underline{60.4} & 99.0 & \underline{64.2} & $^*$95.8 & $^*$74.0 & $^*$52.6& \underline{39.6} & 53.5\\
PR-P2G & 68.8 & 93.0 & 66.7 & 72.8 & 51.0 & 48.6 & 56.9 & 63.9\\
\rowcolor{gray!10}
\hspace{1em} PFER ($\downarrow$) & \wpad9.1 & 12.9 & \wpad6.7 & \wpad7.9 & \wpad5.7 & \wpad3.3 & \wpad6.7 & \wpad6.9\\

\bottomrule
\end{tabular}
}
\caption{WER ($\downarrow$) of different P2G settings on low-resource languages ``Best'' stands for lowest WER in \autoref{tab:main-asr} from ASR models. $^*$ indicates post-processed languages.}
\label{tab:analysis-task-p2g}
\vspace{-2mm}
\end{table}

\paragraph{The language token captures phonotactics}
The language identification (LID) performance of \POWSM on seen languages in FLEURS reaches 92.3\% accuracy, as shown in \autoref{fig:lang-token-seen}.
%Although the model was not explicitly trained for this task (i.e., the language token was never masked out), it achieves 92.3\% accuracy. % Not true. It's always provided, but it's always included in decoder attn loss.
To see if the model implicitly learns phonotactic patterns and associates them with the language token, we evaluate PR on unseen languages by manipulating the language token at inference time. 
For VoxAngeles and Tusom2021, the three most frequently assigned languages are Bashkir (42.6\%, 25.1\%), English (30.2\%, 67.7\%), and Kinyarwanda (14.5\%, 2.3\%), which are all relatively high-resource languages in IPAPack++. 
\autoref{tab:lang-token-unseen} shows that assigning the detected language token yields better performance than always using English, while setting the language as unknown performs best. This indicates that the language token influences PR by shifting the output distribution toward the assigned language.

\begin{table}[h]

\vspace{-2mm}
\centering
\small
\setlength{\tabcolsep}{6pt} % Adjust column spacing
\renewcommand{\arraystretch}{1.2}
%\resizebox{\columnwidth}{!}{
\begin{tabular}{l c c c c c}
\toprule
Lang. token & VoxAngeles & Tusom2021 & Example\\
\midrule
\rowcolor{gray!10}
% \multicolumn{2}{l}{Phonetic transcription}  & &  \ipa{/adʒmɜ/} \\
\multicolumn{2}{l}{Phonetic transcription}  & &  \ipa{/adZm3/} \\
% \texttt{<unk>} & 17.11 & 21.96 & \ipa{/a\textcolor{blue}{d͡ʒ}ima/} \\
\texttt{<unk>} & 17.11 & 21.96 & \textipa{/a\textcolor{blue}{\t{dZ}}ima/} \\
 % Detected & 17.55 & 23.92 & \textipa{/\textcolor{orange}{ɒju}m/}\\
Detected & 17.55 & 23.92 & \textipa{/\textcolor{orange}{6ju}m/}\\
 % \texttt{<eng>} & 19.91 & 24.21 & \ipa{/a\textcolor{orange}{ɪt}imɔ/} \\
 \texttt{<eng>} & 19.91 & 24.21 & \textipa{/a\textcolor{orange}{It}imO/} \\
\bottomrule
  \end{tabular}
 % }
 \caption{PFER ($\downarrow$) of PR performance with different language token. The detected language in the example is \texttt{<bak>}. \textcolor{blue}{Blue} for correct, \textcolor{orange}{orange} for error compared to other examples.}
\label{tab:lang-token-unseen}
\end{table}

\section{Conclusion}

We train a fully open-source phonetic speech foundation model \POWSM using our scalable multi-task framework.
Our model achieves state-of-the-art performance on PR while also supporting ASR across more than 70 languages. 
Beyond PR and ASR, the model's ability to perform audio-guided G2P and P2G enables applications that require fine-grained linguistic analysis such as atypical speech assessment.
Our analysis reveals that \POWSM's encoder benefits from phoneme-level CTC supervision and stronger encoder weighting, enhancing cross-lingual generalization.
Additionally, the model demonstrates interpretable multimodal and language-aware behaviors, effectively mediating between phonetic detail and standardized phonological patterns.
To conclude, \POWSM not only provides a strong phone recognition foundation model for high-resource languages, but also acts as a versatile resource for unseen languages and socio-phonetic variation.

\section{Future Work}
% Skipped: \reviewer{CTC-only model?} \chinjou{vaguely mentioned in limitation; stress again that AED is for multitasking}
% Skipped: \reviewer{tone?} \chinjou{already in limitation}
\POWSM's current decoder serves as a large phoneme-level phonotactic language model on which linguists could investigate hypotheses about phonetic universals \citep{chodroff2024voxangeles,chodroff2025phonetic} and phonotactics \citep{shim2024phonotactic,pimentel2020phonotactic}.
In the future, we seek to adapt to socio-phonetic variation either through (unsupervised) test-time adaptation \citep{lin22b_interspeech}, in-context learning \citep{roll2025context,wang2024can}, or mechanistic interpretability \citep{tang2024language}.
% Mechanistic interpretability techniques could also be used to identify language-agnostic and language-specific neurons \citep{tang2024language} in POWSM to adapt transcriptions to lower-resource languages.
Furthermore, since \citet{shim2025languages} found that earlier encoder layers in Whisper preserve more phonetic detail, early exiting may mitigate the decoder's tendencies to normalize socio-phonetic variation.
% Other future work includes leveraging XEUS, self-supervised speech features learned from 4,000 languages \citep{chen-etal-2024-towards-robust}.
% POWSM will also have forced alignments, by modifying the CTC alignments to be less peaky.
% Finally, we also seek to adapt phoneme-level transcripts to phone-level transcripts to achieve language-universal \textit{phone} recognition. % via reverse phonemicization (see Aim 3 in NSF grant)
% \eunjung{there are lots of future works. while promising, i think we can keep it less, and focus on our contribution and limitations. including future works that are related to limitations will be a good start.}

\section*{Limitations}
\POWSM has several limitations that we aim to address in future work.
First, the model is neither strictly phonemic nor phonetic: its training data consist of cleaned and filtered phonemic transcriptions from multiple languages, which are not fully faithful to the phonetic or phonemic structure of the audio. Although phonemic transcriptions share similarities across languages, adding auxiliary tasks and language tokens may have reinforced language-specific biases. We also currently lack sufficient allophone-level data, which would provide more language-independent information.

Second, the model still favors high-resource languages. Since we include a decoder for language modeling and language tokens, both of which function effectively, the model would inherently bias toward the seen distribution.

Finally, the current AED architecture, although effective for multitasking, imposes certain engineering limitations. Inference is significantly slower than with encoder-only models, and the architecture does not easily support tone modeling, limiting its application to tonal languages.

\subsubsection*{Ethics Statement}

All of our data is ethically sourced, either through permissive licensing or through proper consent.
We are aware of the implicit prescriptivism and representational harms \citep{crawford} that normalizing socio-phonetic variation in ASR or PR models can create.
This may threaten linguistic diversity instead of preserving it.
% ``We therefore call on the speech community to start preserving sociolinguistic variation at the same time that it improves WER for all demographics.'' kalvin's interspeech paper notes (never made it to the final version)
We also acknowledge that accurate modeling of socio-phonetic variation can enable demographic inference, as demographics and phonetic variation are deeply intertwined \citep{labov1963social}.
We stress that uses of \POWSM must align with our vision: a future where advances in spoken language processing and NLP do not leave low-resource varieties behind.

\subsubsection*{The Use of LLMs}
We acknowledge the use of large language models (LLMs) to assist with grammar correction and clarity improvements in writing this paper. All conceptual, methodological, and experimental contributions were developed independently by the authors.

% \iffalse

% \kalvin{reminder to take out just for peer review}
\section*{Acknowledgments}
This work used Bridges2 in the PSC and Delta NCSA computing systems through allocation CIS210027 from the Advanced Cyberinfrastructure Coordination Ecosystem: Services \& Support (ACCESS) program, supported by National Science Foundation grants.
% \shinji{TODO}.
We appreciate the contributions of Kevin Glocker, Shih-heng Wang, Farhan Samir, and Aaricia Herygers during earlier iterations of this work. 
We are also grateful for feedback from David Harwath, Brendon Boldt, Sanjay Subramanian, Rudy Corona, Anya Ji, Seun Eisape, and Kayo Yin.

% \fi

\bibliography{custom,doreco}

\newpage
\appendix

\section{Appendix}

\subsection{Refining English G2P}
\label{sec:appendix-engg2p}
We observed confusion in plosive voice-onset times on unseen languages in preliminary experiments, which is likely from English G2P data. For instance, broad phonemic transcription in English typically uses /b/ to transcribe the /b/ in /bat/, but its voice onset timing is actually voiceless in Mainstream American English and is closer to [p]. To mitigate this, we apply rule-based refinements to English G2P transcriptions, adjusting plosive voicing and aspiration, lateral velarization, and vowel nasalization.

The rules are listed below: 
1) word-initial voiceless plosives (\textipa{/p/, /t/, /k/}) are aspirated,
2) word-initial voiced plosives (\textipa{/b/, /d/, /g/}) are voiceless, 
3) lateral \textipa{/l/} is velarized at the end of syllables, and 
4) vowel nasalization before nasal consonants.

\subsection{Baseline Implementation}
\label{sec:appendix-baseline}
We provide the baselines' training data source, number of languages covered in the data, and links to model checkpoints or repository in \autoref{tab:training-data-summary}.

\begin{table*}[t]
\centering
\small
\resizebox{\textwidth}{!}{
%\begin{tabular}{@{}l p{4cm} c p{5cm}}
\begin{tabular}{ll c l}
\toprule
\textbf{Model} & \textbf{Training Data Sources} & \textbf{Language Coverage} & \textbf{Model checkpoint / GitHub} \\
\midrule
\textbf{PR baselines}\\
\addlinespace[0.10cm]

\textbf{Allosaurus} & VoxForge,  Japanese CSJ, Hkust & \multirow{2}{*}{12} & \href{https://github.com/xinjli/allosaurus}{xinjli/allosaurus}  \\
\cite{li2020universal,li2021hierarchical} & Tedlium, Switchboard etc & & \\
\addlinespace[0.15cm]

\textbf{Allophant} &\multirow{2}{*}{Common Voice 10.0} & \multirow{2}{*}{34} & \href{https://github.com/kgnlp/allophant}{kgnlp/allophant}  \\
\cite{glocker23_interspeech} & & &  \\
\addlinespace[0.15cm]

\textbf{Wav2Vec2Phoneme} & MLS, Common Voice, & \multirow{2}{*}{40+} &\multirow{2}{*}{{\href{https://huggingface.co/facebook/wav2vec2-xlsr-53-espeak-cv-ft}{facebook/wav2vec2-xlsr-53-espeak-cv-ft}}} \\
\cite{xu22b_interspeech} & Babel &  \\
\addlinespace[0.15cm]

\textbf{MultIPA} & \multirow{2}{*}{Common Voice 11.0} & \multirow{2}{*}{7} & \multirow{2}{*}{{\href{https://huggingface.co/ctaguchi/wav2vec2-large-xlsr-japlmthufielta-ipa1000-ns}{ctaguchi/wav2vec2-large-xlsr-japlmthufielta-ipa1000-ns}}} \\
\cite{taguchi23_interspeech} &\\
\addlinespace[0.15cm]

\textbf{ZIPA} & IPAPack++ & \multirow{2}{*}{88} & \href{https://github.com/lingjzhu/zipa}{lingjzhu/zipa} \\
\cite{zhu-etal-2025-zipa} & MMS ulab v2.,  VoxLingua-107 (Pseudo-label) & & \href{https://huggingface.co/anyspeech/zipa-large-crctc-ns-800k}{anyspeech/zipa-large-crctc-ns-800k} \\
\addlinespace[0.10cm]

\midrule
\textbf{ASR baselines}\\
\addlinespace[0.10cm]

\textbf{OWSM-CTC v4} & \multirow{2}{*}{OWSM v3.2, YODAS} & \multirow{2}{*}{100+} &\multirow{2}{*}{{\href{https://huggingface.co/espnet/owsm_ctc_v4_1B}{espnet/owsm\_ctc\_v4\_1B}}}\\
\cite{peng25c_interspeech} \\
\addlinespace[0.15cm]

\textbf{OWLS} & \multirow{2}{*}{OWSM v3.2, YODAS} & \multirow{2}{*}{150} &\multirow{2}{*}{{\href{https://huggingface.co/collections/espnet/owls-scaling-laws-for-speech-recognition-and-translation-67ab7f991c194065f057ce8d}{espnet/owls-scaling-laws-for-speech-recognition-and-translation}}}\\
\cite{chen2025owls}\\

\bottomrule
\end{tabular}
}
\caption{Overview of the baselines for our work.}
\label{tab:training-data-summary}
\end{table*}

\subsection{FLEURS language selection for ASR}
\label{sec:appendix-asr-splits}
We first filter out languages with more than 8 hours of training data in IPAPack++ \cite{zhu-etal-2025-zipa}, keeping only those that are also present in FLEURS.
Then, following the training data amounts reported in \citet{chen2025owls}, we further identify the 50 lowest-resource languages to exclude any that may have other substantial sources not included in IPAPack++.
This process leaves us with nine languages. We finally exclude \texttt{ell}, as it is comparatively higher-resource and because there are already three other Balto-Slavic languages.
Note that other models use strictly more data than ours---not only in terms of dataset count but also because IPAPack++ applies additional data-quality filtering. \autoref{tab:asr-data-stats} lists the amount of ASR training data for baselines.

\begin{table}[h]
\vspace{-2mm}
\centering
\small
\setlength{\tabcolsep}{4pt} % Adjust column spacing
\renewcommand{\arraystretch}{1.4}
\resizebox{\columnwidth}{!}{
\begin{tabular}{l cc c cc ccc}
\toprule
% use numbers reported in OWSM v4
& \multicolumn{2}{c}{Afroasiatic} & Turkic & \multicolumn{2}{c}{Indo-Iranian} & \multicolumn{3}{c}{Balto-Slavic}\\
Model & \texttt{afr} & \texttt{orm} & \texttt{aze}  & \texttt{pan} & \texttt{tgk}  & \texttt{mkd} & \texttt{bos} & \texttt{slv}\\
\midrule
POWSM & 2.71 & 5.11 & 6.89 & 4.96 & 6.52 & 5.14 & 7.57 & 7.19 \\
OWSM-CTC v4 & \multirow{2}{*}{5.54} & \multirow{2}{*}{6.50} & \multirow{2}{*}{10.69} & \multirow{2}{*}{8.30} & \multirow{2}{*}{8.03} & \multirow{2}{*}{8.4} & \multirow{2}{*}{9.96} & \multirow{2}{*}{26.00} \\
OWLS & \\
\bottomrule
  \end{tabular}
}
 \caption{Amount of ASR training data for languages included in ASR comparison (in hours), according to \citet{zhu-etal-2025-zipa} and \citet{chen2025owls}.}
 \label{tab:asr-data-stats}
% \vspace{-2mm}
\end{table}

\begin{figure*}
    \centering
    \includegraphics[width=0.8\textwidth]{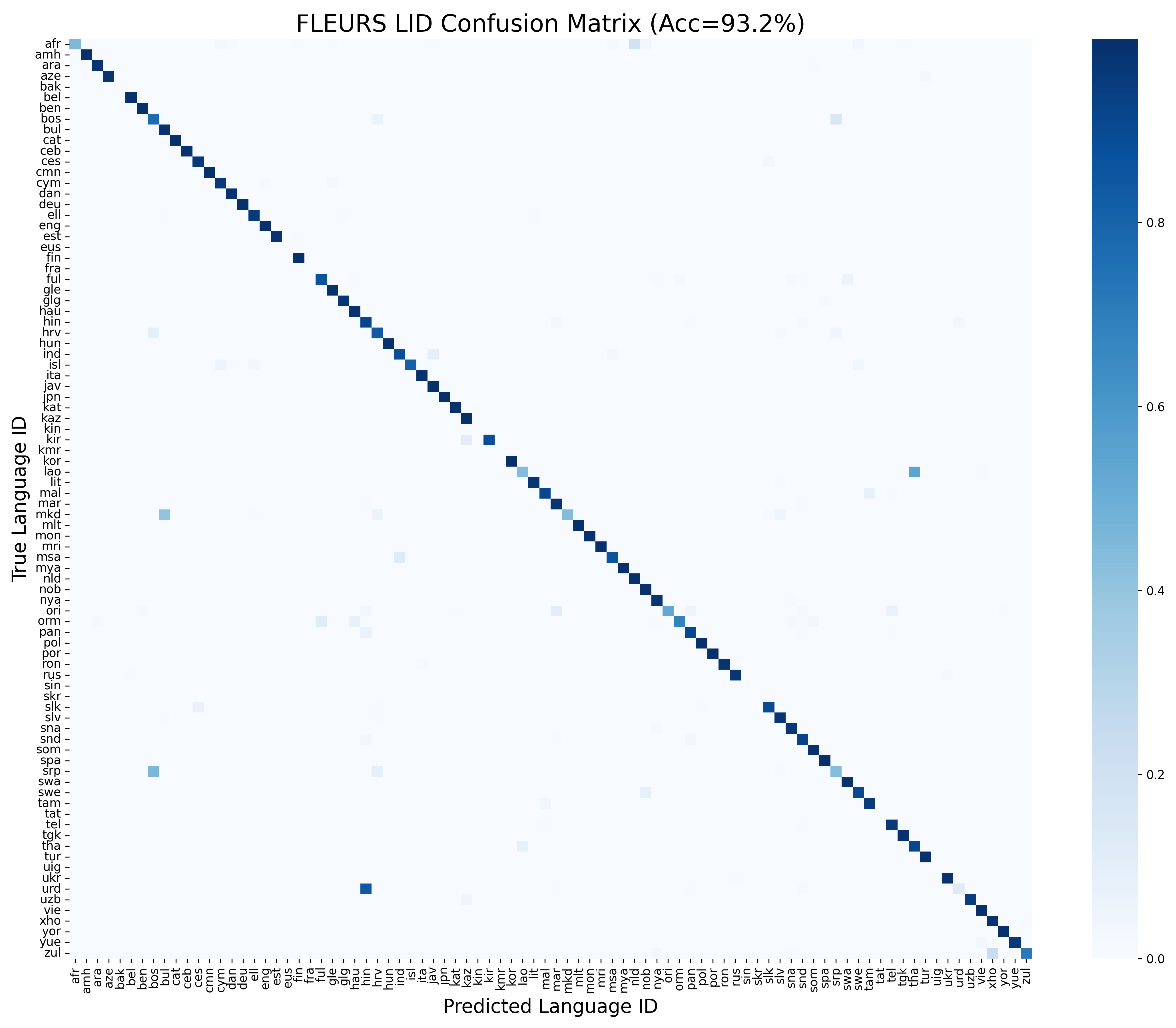}
    \caption{Confusion matrix of LID on FLEURS.}
    \label{fig:lang-token-seen}
\end{figure*}

\subsection{ASR Performance on In-Domain Data}
\label{sec:appendix-asr-perf}
We run greedy decoding (\texttt{ctc=0.0, beam=1}) on the test sets of in-domain data. 
\autoref{tab:asr-hrl} shows that ASR performance is reasonably competitive with previous models \cite{radford2023robust, peng25c_interspeech} of similar size and architecture despite being trained on much fewer data. 
The performance gap is smaller for languages that constitute a larger portion of the training data, and larger for mid-frequency languages.
Errors often involve substitutions of phonetically similar words or phrases (e.g., \textit{``mostly rapscallions as fur''} to \textit{``mostly wrapped skaggins as fer''}).

\begin{table}[h]
\centering
\small
\setlength{\tabcolsep}{4pt} % Adjust column spacing
\renewcommand{\arraystretch}{1.2}
\resizebox{\columnwidth}{!}{
\begin{tabular}{l ccc ccc ccc cc}
\toprule
& \texttt{eng} & \texttt{deu} & \texttt{nld} & \texttt{fra} & \texttt{ita} & \texttt{spa} & \texttt{por} & \texttt{pol} & \texttt{tam} & \texttt{kaz} & \texttt{cmn} \\
\midrule 
\POWSM & --- & 11.5 & 20.2 & 13.8 & 23.6 & \wpad9.4 & 25.7 & 31.0 & 40.4 & 23.4 & 21.9 \\
\POWSM-fix & 12.2 & 12.7 & 19.5 & 13.9 & 22.2 & 11.3 & 25.5 & 28.5 & 32.6 & 14.8 & 21.7 \\
Whisper & \wpad8.3 & 11.5 & 18.2 & 13.6 & 21.3 & \wpad9.1 & 13.8 & \textbf{12.5} & --- & --- & 26.6 \\
OWSM & \textbf{\wpad4.7} & 10.3 & 15.7 & 14.1 & 21.8 & \wpad7.1 & 11.5 & 16.0 & --- & --- & \textbf{14.9}\\
\midrule
OWSM-CTC & \wpad5.1 & \textbf{\wpad9.5} & \textbf{15.1} & \textbf{\wpad7.8} & \textbf{15.5} & \textbf{\wpad5.8} & \textbf{10.3} & 15.1 & --- & --- & 18.7\\
\bottomrule
  \end{tabular}
 }
\caption{WER ($\downarrow$) on test sets; \texttt{cmn} reports CER ($\downarrow$). \textit{POWSM-fix} is discussed in \autoref{sec:appendix-textissue}. 
\textit{Whisper} (Whisper-small) and \textit{OWSM} (OWSM v4 small) are models of similar size and architecture, trained with 10 times more data. \texttt{tam} and \texttt{kaz} are omitted due to normalization issues; Whisper outputs numerals for \texttt{cmn}, inflating its CER. }
\label{tab:asr-hrl}
\end{table}

\subsection{Multi-tasking at Different Scales}
\label{sec:appendix-multitask}
Multi-tasking may improve performance by tying acoustic signals to well-defined symbolic representations, yet it may distract the model if the relationships are not learned effectively. 
We train \POWSM with different data and model scales to examine how multitask learning interacts with the setup, and use \texttt{beam=1} during decoding to speed up inference.

\autoref{tab:multitask-small} shows that there is no clear trend regarding whether multitasking benefits PR performance. PR performance  degrades when the model has excessive capacity relative to the available data (too little data), or when it is limited by size (too much data).

Further evidence is needed before concluding that phoneme recognition benefits less from scaling, as we currently lack sufficient data and large model capacity to test this thoroughly. Nevertheless, the model demonstrates the ability to multitask, which represents a promising direction for future work. 

% \autoref{tab:multitask-small} shows that in a low-resource setting, multi-tasking consistently improves performance as more tasks are added. However, the gains saturate with larger data and models, echoing prior findings \citep{chen2025owls} that less language-dependent tasks such as phone recognition benefit less from scaling.
% Our results contradict ``How Phonotactics Affect Multilingual and Zero-Shot ASR Performance''
% ``We show that the gain from modeling crosslingual phonotactics is limited, and imposing a too strong model can hurt the zero-shot transfer. Furthermore, we find that a multilingual LM hurts a multilingual ASR system’s performance, and retaining only the target language’s phonotactic data in LM training is preferable.''

\begin{table}[h]
\centering
\small
\setlength{\tabcolsep}{4pt} % Adjust column spacing
\renewcommand{\arraystretch}{1.2}
\resizebox{\columnwidth}{!}{
\begin{tabular}{c c c c c c}
\toprule
Data ( khr) & Tasks & Params. & VoxAngeles & Tusom2021 & L2-Arctic \\
\midrule 
0.25 & 1 & 100M & 27.22 & 32.59 & 13.50\\
0.25 & 4 & 100M & 26.30 & 30.32 & 13.40\\
0.25 & 1 & 300M & 20.63 & 25.83 & 12.88\\
%0.25 & 2 & 300M & 20.10 & 27.53 & \\
0.25 & 4 & 300M & 23.81 & 25.91 & 14.14\\
\midrule
17 & 1 & 100M & 17.88 & 26.68 & 11.76\\
17 & 2 & 100M & 24.69 & 49.28 & 11.35\\
17 & 4 & 100M & 30.07 & 61.89 & 12.37\\
17 & 1 & 300M & 17.08 & 25.20 & 10.50\\
17 & 2 & 300M & 17.17 & 23.70 & 10.47\\
17 & 4 & 300M & 17.58 & 33.52 & 10.54\\
\bottomrule
  \end{tabular}
 }
\caption{Comparison of PFER ($\downarrow$) on different seting of tasks and data. 1 task refers to PR, 2 tasks refer to PR+ASR, and 4 tasks include PR, ASR, P2G, and G2P.}
\label{tab:multitask-small}
\vspace{-2mm}
\end{table}

\subsection{DoReCo \citep{paschen2020doreco}}
\label{sec:appendix-doreco}

DoReCo consists of the following constituent datasets: \citep{
doreco-anal1239,
doreco-apah1238,
doreco-arap1274,
doreco-bain1259,
doreco-beja1238,
doreco-bora1263,
doreco-cabe1245,
doreco-cash1254,
doreco-dolg1241,
doreco-even1259,
doreco-goem1240,
doreco-goro1270,
doreco-hoch1243,
doreco-jeha1242,
doreco-jeju1234,
doreco-kaka1265,
doreco-kama1351,
doreco-kark1256,
doreco-komn1238,
doreco-ligh1234,
doreco-lowe1385,
doreco-movi1243,
doreco-ngal1292,
doreco-nisv1234,
doreco-nngg1234,
doreco-nort2641,
doreco-nort2875,
doreco-orko1234,
doreco-pnar1238,
doreco-port1286,
doreco-resi1247,
doreco-ruul1235,
doreco-sadu1234,
doreco-sanz1248,
doreco-savo1255,
doreco-sout2856,
doreco-sout3282,
doreco-stan1290,
doreco-sumi1235,
doreco-svan1243,
doreco-taba1259,
doreco-teop1238,
doreco-texi1237,
doreco-trin1278,
doreco-tsim1256,
doreco-urum1249,
doreco-vera1241,
doreco-warl1254,
doreco-yong1270,
doreco-yuca1254,
doreco-yura1255}.

\subsection{Fixing ASR Text Normalization}
\label{sec:appendix-textissue}
After submission, we discovered a text normalization issue specific to Librispeech that degraded its ASR performance. 
Continuing training with the corrected data for 20 GPUs hours on H100s mostly solved the problem (WER=14.2). 

Additionally, we trained a model from scratch with updated data, denoted as \textit{POWSM-fix}, where ASR transcripts are lowercased and punctuation is removed except for apostrophes and hyphens.
As shown in \autoref{tab:asr-hrl}, ASR performance is similar or improved, while \autoref{tab:powsm-fix} shows comparable PR performance on out-of-domain data when decoding with \texttt{ctc=0.3, beam=1}, enabling faster inference.
Both checkpoints are publicly available on HuggingFace for reproducibility.

\begin{table}[h]
\centering
\small
\setlength{\tabcolsep}{2pt} % Adjust column spacing
\renewcommand{\arraystretch}{1.2}
\resizebox{\columnwidth}{!}{
\begin{tabular}{l ccc ccccc}
\toprule
& DoReCo & VoxA. & Tusom. & Buckeye & DRC-SE & L2-ARC & EpaDB & SO762 \\
\midrule 
\POWSM & 17.06 & 17.11 & 21.96 & 12.63 & 18.33 & 11.32 & 11.86 & 17.84 \\
\POWSM-fix & 19.06 & 18.80 & 22.73 & 13.22 & 18.84 & 10.94 & 10.98 & 16.56 \\ % ctc=0.3, beam=1
% \POWSM-fix & --- & 19.05 & 24.38 & 12.87 & 19.18 & 12.15 & 10.59 & 16.03 \\ % ctc=0.3, beam=3
\bottomrule
  \end{tabular}
 }
\caption{PFER ($\downarrow$) comparison on out-of-domain data.}
\label{tab:powsm-fix}
\end{table}

\end{document}